\documentclass[letterpaper]{article}
\usepackage{aaai}
\usepackage{times}
\usepackage{helvet}
\usepackage{courier}
\usepackage{amsmath, xparse}
\usepackage{rotating}

\usepackage{hyperref}
\usepackage{cleveref}
\usepackage{multirow}
\usepackage{amsfonts}
\usepackage{booktabs}
\usepackage{footmisc}

\usepackage{xcolor}
\usepackage{graphicx}
\graphicspath{ {./graphics/} }

\crefformat{section}{\S#2#1#3} 
\crefformat{subsection}{\S#2#1#3}
\crefformat{subsubsection}{\S#2#1#3}

\usepackage{textcomp}  
\usepackage{scalerel}  

\newcommand{\libname}{OpenHands}

\def\libfull{\scalerel*{\includegraphics{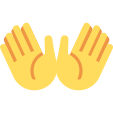}}{\textrm{\textbigcircle}} \texttt{OpenHands}}

\begin{document}



\nocopyright
\title{\libfull: Making Sign Language Recognition Accessible with Pose-based Pretrained Models across Languages}
\author {
    Prem Selvaraj\thanks{Equal contribution.}\textsuperscript{\rm 1},
    Gokul NC\footnotemark[1]\textsuperscript{\rm 1},
    Pratyush Kumar\textsuperscript{\rm 1,2,3},
    Mitesh Khapra\textsuperscript{\rm 1,2}\\
    \textsuperscript{\rm 1}AI4Bharat,
    \textsuperscript{\rm 2}IIT-Madras,
    \textsuperscript{\rm 3}Microsoft Research\\
    prem@ai4bharat.org,
    gokulnc@ai4bharat.org,
    pratyush@cse.iitm.ac.in,
    miteshk@cse.iitm.ac.in
}

\maketitle
\begin{abstract}
    AI technologies for Natural Languages have made tremendous progress recently. 
    However, commensurate progress has not been made on Sign Languages, in particular, in recognizing signs as individual words or as complete sentences.
    We introduce \libfull\footnote{\url{https://github.com/AI4Bharat/OpenHands}}, a library where we take four key ideas from the NLP community for low-resource languages and apply them to sign languages for word-level recognition.
    First, we propose using pose extracted through pretrained models as the standard modality of data to reduce training time and enable efficient inference, and
    we release standardized pose datasets for 6 different sign languages - American, Argentinian, Chinese, Greek, Indian, and Turkish.
    Second, we train and release checkpoints of 4 pose-based isolated sign language recognition models across all 6 languages, providing baselines and ready checkpoints for deployment.
    Third, to address the lack of labelled data, we propose self-supervised pretraining on unlabelled data. 
    We curate and release the largest pose-based pretraining dataset on Indian Sign Language (Indian-SL). 
    Fourth, we compare different pretraining strategies and for the first time establish that pretraining is effective for sign language recognition by demonstrating (a) improved fine-tuning performance especially in low-resource settings, and (b) high crosslingual transfer from Indian-SL to few other sign languages.
    We open-source all models and datasets in \libfull~with a hope that it makes research in sign languages more accessible.
\end{abstract}

\section{Introduction}\label{sec:introduction}

According to the World Federation of the Deaf, there are approximately 72 million Deaf people worldwide. More than 80\% of them live in developing countries. Collectively, they use more than 300 different sign languages varying across different nations \cite{UN_SL_Day}. Loss of hearing severely limits the ability of the Deaf to communicate and thereby adversely impacts their quality of life. 
In the current increasingly digital world, systems to ease digital communication between Deaf and hearing people are important accessibility aids.
AI has a crucial role to play in enabling this accessibility with automated tools for Sign Language Recognition (SLR).
Specifically, transcription of sign language as complete sentences is referred to as Continuous Sign Language Recognition (CSLR), while recognition of individual signs is referred to as Isolated Sign Language Recognition (ISLR).
There have been various efforts to build datasets and models for ISLR and CLSR tasks \cite{Adaloglou_2021,koller2020quantitative}. 
But these results are often concentrated on a few sign languages (such as the American Sign Language) and are reported across different research communities with few standardized baselines.
When compared against text- and speech-based NLP research, the progress in AI research for sign languages is significantly lagging. 
This lag has been recently brought to notice of the wider NLP community \cite{yin2021including}. 


For most sign languages across the world, the amount of labelled data is very low and hence they can be considered \textit{low-resource languages}. 
In the NLP literature, many successful templates have been proposed for such low-resource languages. 
In this work, we adopt and combine many of these ideas from NLP to sign language research. 
We implement these ideas and release several datasets and models in an open-source library \libfull~with the following key contributions:

\noindent\textbf{1. Standardizing on pose as the modality:}
For natural language understanding (NLU) tasks, such as sentiment classification, it is standard to use a pretrained encoder, such as BERT.
This task-agnostic encoder significantly reduces need for labelled data on the NLU task. 
Similarly for SLR tasks, we propose to standardize on a pose-extractor as an encoder, which processes raw RGB videos and extracts the frame-wise coordinates for few keypoints. 
Pose-extractors are useful across sign languages and also other tasks such as action recognition \cite{AAAI1817135,liu2020disentangling}, and can be trained to high accuracy.
Further, as we report, pose as a modality makes both training and inference for SLR tasks efficient. 
We release pose-based versions of existing datasets for 6 sign languages: American, Argentinian, Chinese, Greek, Indian, and Turkish.

\noindent\textbf{2. Standardized comparison of models across languages:} 
The progress in NLP has been earmarked by the release of standard datasets, including multilingual datasets like XGLUE \cite{liang2020xglue}, on which various models are compared.
As a step towards such standardization for ISLR, we train 4 different models spanning sequence models (LSTM and Transformer) and graph-based models (ST-GCN and SL-GCN) on 7 different datasets for sign languages mentioned above, and compare them against models proposed in the literature. 
We release all 28 trained models along with scripts for efficient deployment which demonstrably achieve real-time performance on CPUs and GPUs.

\noindent\textbf{3. Corpus for self-supervised training:} 
A defining success in NLP has been the use of self-supervised training, for instance masked-language modelling \cite{devlin2018bert}, on large corpora of natural language text.
To apply this idea to SLR, we need similarly large corpora of sign language data.
To this end, we curate 1,129 hours of video data on Indian Sign Language.
We pre-process these videos with a custom pipeline and extract keypoints for all frames. 
We release this corpus which is the first such large-scale sign language corpus for self-supervised training. 

\noindent\textbf{4. Effectiveness of self-supervised training:} 
Self-supervised training has been demonstrated to be effective for NLP: Pretrained models require small amounts of fine-tuning data \cite{devlin2018bert,baevski2020wav2vec} and multilingual pretraining allows crosslingual generalization \cite{hu2020xtreme}.
To apply this for SLR, we evaluate multiple strategies for self-supervised pretraining of ISLR models and identify those that are effective. 
With the identified pretraining strategies, we demonstrate the significance of pretraining by showing improved fine-tuning performance, especially in very low-resource settings and also show high crosslingual transfer from Indian SL to other sign languages. 
This is the first and successful attempt that establishes the effectiveness of self-supervised learning in SLR.
We release the pretrained model and the fine-tuned models for 4 different sign languages.



Through these datasets, models, and experiments we make several observations. 
First, in comparing standardized models across different sign languages, we find that graph-based models working on pose modality define state-of-the-art results on most sign languages. 
LSTM-based models lag on accuracy but are significantly faster and thus appropriate for constrained devices.
Second, we firmly establish that self-supervised pretraining helps as it improves on equivalent models trained from scratch on labelled ISLR data.
The performance gap is particularly high if the labelled data contains fewer samples per label, i.e., for the many sign languages which have limited resources the value of self-supervised pretraining is particularly high.
Third, we establish that self-supervision in one sign language (Indian SL) can be crosslingually transferred to improve SLR on other sign languages (American, Chinese, and Argentinian).
This is particularly encouraging for the long tail of over 300 sign languages that are used across the globe.
Fourth, we establish that for real-time applications, pose-based modality is preferable over other modalities such as RGB, use of depth sensors, etc.~due to reduced infrastructure requirements (only camera), and higher efficiency in self-supervised pretraining, fine-tuning on ISLR, and inference.
We believe such standardization can help accelerate dataset collection and model benchmarking.
Fifth, we observe that the trained checkpoints of the pose-based models can be directly integrated with pose estimation models to create a pipeline that can provide real-time inference even on CPUs.
Such a pipeline can enable the deployment of these models in real-time video conferencing tools, perhaps even on smartphones.

As mentioned all datasets and models are released with permissible licenses in \libfull~with the intention to make SLR research more accessible and standardized.
We hope that others contribute datasets and models to the library, especially representing the diversity of sign languages used across the globe.

The rest of the paper is organized as follows. In \cref{sec:background} we present a brief overview of the existing work. In \cref{sec:standardization} we describe our efforts in standardizing datasets and models across six different sign languages. In \cref{sec:ssl_islr} we explain our pretraining corpus and strategies for self-supervised learning and detail results that establish its effectiveness.
In \cref{sec:slr_library} we describe in brief the functionalities of the \libfull~library.
In \cref{sec:future_works}, we summarize our work and also list potential follow-up work.

\section{Background and Related Work}\label{sec:background}

Significant progress has been made in Isolated Sign Language Recognition (ISLR) due to the release of datasets \cite{li2020word,Sincan_2020,chai2014devisign,8466903} and recent deep learning architectures \cite{Adaloglou_2021}. This section reviews this work, with a focus on pose-based models.

\subsection{Sign Language}

A sign language (SL) is the visual language used by the Deaf and hard-of-hearing (DHH) individuals, which involves usage of various bodily actions, like hand gestures and facial expressions, called signs to communicate.
A sequence of signs constitutes a phrase or sentence in a SL.
The signs can be transcribed into sign-words of any specific spoken language usually written completely in capital letters.
Each such sign-word is technically called as a gloss and is the basic atomic token of an SL transcript.




The task of converting each visual sign communicated by a signer into a gloss is called isolated sign language recognition (ISLR).
The task of converting a continuous sequence of visual signs into serialized glosses is referred as continuous sign language recognition (CSLR).
CSLR can either be modeled as an end-to-end task, or as a combination of sign language segmentation and ISLR.
The task of converting signs into spoken language text is referred as sign language translation (SLT), which can again either be end-to-end or a combination of CLSR and gloss-sequence to spoken phrase converter.

Although SL content is predominantly recorded as RGB (color) videos, it can also be captured using various other modalities like depth maps or point cloud, finger gestures recorded using sensors, skeleton representation of the signer, etc.
In this work, we focus on ISLR using pose-skeleton modality.
A pose representation, extracted using pose estimation models, provides the spatial coordinates at which the joints such as elbows and knees, called keypoints, are located in a field or video.
This pose information can be represented as a connected graph with nodes representing keypoints and edges may be constructed across nodes to approximately represent the human skeleton. 

For ISLR, since it is generally modeled as a single-label classification problem, the de-facto metric to measure performance and quality of a model is \textit{accuracy} (or \textit{top-1 accuracy}), although other top-k accuracies can also be reported.
Usually while building an ISLR dataset, it is important to ensure that there are enough diverse samples per class, with each sample being curated from different signers, environment, and other factors like camera-orientations, pace of signing, etc.
This ensures that the whole dataset can be split in such a way that the training distribution isn't significantly similar to the validation and test sets, inorder to be able to build models that generalize to real-world scenarios.

\subsection{Models for ISLR}\label{sec:models_survey}

Initial methods for SLR focused on hand gestures from either video frames \cite{isl_hand_2020} or sensor data such as from smart gloves \cite{glove_talk}.
Given that such sensors are not commonplace and that body posture and face expressions are also of non-trivial importance for understanding signs \cite{Hu2020GlobalLocalEN}, convolutional network based models have been used for SLR \cite{8316344}. 

The ISLR task is related to the more widely studied action recognition task \cite{zhu2020comprehensive}. 
Like in action recognition task, highly accurate pose recognition models like OpenPose \cite{cao2018openpose} and MediaPipe Holistic \cite{holistic:online} are being used for ISLR models \cite{li2020word,10.1145/3264746.3264805}, where frame-wise keypoints are the inputs.
Although RGB-based models may outperform pose-based models \cite{li2020word} narrowly, pose-based models have far fewer parameters and are more efficient for deployment if used with very-fast pose estimation pipelines like MediaPipe.
In this work, we focus on lightweight pose-based ISLR which encode the pose frames and classify the pose using specific decoders.
We briefly discuss the two broad types of such models: sequence-based and graph-based.



Sequence-based models process data sequentially along time either on one or both directions. 
Initially, RNNs were used for pose-based action recognition to learn from temporal features \cite{7298714,Zhang_2017,si2018skeletonbased}.
Specifically, sequence of pose frames are input to GRU or LSTM layers, and the output from the final timestep is used for classification.
Transformer architectures with encoder-only models like BERT \cite{vaswani2017attention} have also been studied for pose-based ISLR models \cite{de-coster-etal-2020-sign}.
The input is a sequence of pose frames along with positional embeddings. 
A special [CLS] token is prepended to the sequence, whose final embedding is used for classification.

Graph convolution networks \cite{kipf2017semisupervised}, which are good at modeling graph data have been used for skeleton action recognition to achieve state-of-the-art results, by considering human skeleton sequences as spatio-temporal graphs \cite{cheng2020eccv,liu2020disentangling}.
Spatial-Temporal GCN (ST-GCN) uses human body joint connections for spatial connections and temporal connections across frames to construct a 3d graph, which is processed by a combination of spatial graph convolutions and temporal convolutions to efficiently model the spatio-temporal data \cite{Lin_2020}. 
Many architectural improvements have been proposed over ST-GCN for skeleton action recognition \cite{Zhang_2020,2sagcn2019cvpr,8954160,Cheng_2020_CVPR,cheng2020eccv,liu2020disentangling}.
MS-AAGCN \cite{Shi_2020} uses attention to adaptively learn the graph topology and also proposes STC-attention module to adaptively  weight joints, frames and channels.
Decoupled GCN \cite{cheng2020eccv} improves the capacity of ST-GCN without adding additional computations and also proposes attention guided drop mechanism called DropGraph as a regularization technique.
Sign-Language GCN (SL-GCN) \cite{jiang2021skeleton} combines STC-attention with Decoupled-GCN and extends it to ISLR achieving state-of-the-art results.

\subsection{Pretraining strategies}\label{sec:pretraining_strategies}
We now survey three broad classes of pretraining strategies that we reckon could be applied to SLR.

\subsubsection{Masking-based pretraining}
In NLP, masked language modelling is a pretraining technique where randomly masked tokens in the input are predicted. 
This approach has been explored for action recognition \cite{motion_transformer}, where certain frames are masked and a regression task estimates coordinates of keypoints.
In addition, a direction loss is also proposed to classify the quadrant where the motion vector lies.


\subsubsection{Contrastive-learning based}
Contrastive learning is used to learn feature representations of the input to maximize the agreement between augmented views of the data \cite{contrastive_gao21a,li2021crossclr}. 
For positive examples, different augmentations of the same data item are used, while for negative samples randomly-chosen data items usually from a few last training batches are used.
A variant of contrastive loss called InfoNCE \cite{oord2018representation} is used to minimize the distance between positive samples.

\subsubsection{Predictive Coding}
Predictive Coding aims to learn data representation by continuously correcting its predictions about data in future timesteps given data in certain input timesteps. 
Specifically, the training objective is to pick the future timestep's representation from other negative samples which are usually picked from recent previous timesteps of the same video.
This technique was explored for action recognition in a model called Dense Predictive Coding (DPC) \cite{han2019video}.
Instead of predicting at the frame-level, DPC introduces coarse-prediction at the scale of non-overlapping windows.


Specifically, instead of passing each frame to the model and trying to learn the representations at jittery fine-grained level, the input is partitioned into temporally coarse-grained form by using consecutive non-overlapping windows of equal length.
The encoder produces embeddings for each window.
Finally, for $W_n$ windows, $W_n-W_p$ windows are used as input and $W_p$ windows are used to predict the subsequent future representations using recurrent neural networks.

To pretrain the model, a loss function based on InfoNCE (similar to contrastive learning) is used \cite{mikolov2013distributed,oord2018representation}. Given the estimated future state representations \{\({z}_{t}\), \({z}_{t+1}\), ... , \({z}_{end}\)\} and corresponding actual (encoded) representations \{\(\hat{z}_{t}\), \(\hat{z}_{t+1}\), ... , \(\hat{z}_{end}\)\},  the loss function is constructed as:
\begin{equation}
\mathcal{L}=-\sum_{i}\Bigg[\log{\frac{ \exp(\hat{z}_{i} \cdot {z}_{i})}
	{\sum_{j}\exp(\hat{z}_{i} \cdot {z}_{j})}}\Bigg] 
\end{equation}\label{eq:loss}


\begin{table*}[bt]
\centering
\begin{tabular}{llllllll} \toprule
\textbf{Dataset} & \textbf{Language} &  \textbf{Vocab} & \textbf{Signers} & \textbf{Videos} & \textbf{Hrs} & \textbf{Data} \\ \midrule

AUTSL \cite{Sincan_2020} & Turkish & 226 & 43 & 38,336 & 20.5 & RGBD \\ 
CSL \cite{8466903} & Chinese & 100 & 5 & 500 & 108.84 & RGBD \\ 
WLASL \cite{li2020word} & American & 2000 & 119 & 21,083 & 14 & RGB \\ 
GSL \cite{Adaloglou_2021} & Greek & 310 & 7 & 40785 & 6.44 & RGBD \\ 
LSA64 \cite{Ronchetti2016} & Argentinian & 64 & 10 & 3,200 & 1.90 & RGB \\ 
DEVISIGN \cite{chai2014devisign} & Chinese & 4414 & 30 & 331050 & 21.87 & RGBD \\ 
INCLUDE \cite{include_isl} & Indian & 263 & 7 & 4,287 & 3.57 & RGB \\ \bottomrule

\end{tabular}
\caption{The diverse set of existing ISLR datasets which we study in this work through pose-based models}
\label{table:datasets_list}
\end{table*}

\section{Standardized Pose-based ISLR Models across Sign Languages}\label{sec:standardization}
In this section we describe our efforts to curate standardized pose-based datasets across multiple sign languages and benchmark multiple ISLR models on them.


\subsection{ISLR Datasets}

Multiple datasets have been created for the ISLR task across sign languages. 
However, the amount of data significantly varies across different sign languages, with American and Chinese having the largest datasets currently.
With a view to cover a diverse set of languages, we study 7 different datasets across 6 sign languages as summarised in \Cref{table:datasets_list}. 
For each of these datasets, we generate pose-based data using the Mediapipe pose-estimation pipeline \cite{holistic:online}, which enables real-time inference in comparison with models such as OpenPose \cite{cao2018openpose}.
Mediapipe, in our chosen Holistic mode, returns 3d coordinates for 75 keypoints (excluding the face mesh).
Out of these, we select only 27 sparse 2d keypoints which convey maximum information, covering upper-body, hands and face.
Thus, each input video is encoded into a vector of size $F\times K \times D$, where $F$ is the number of frames in the video, $K$ is the number of keypoints (27 in our case), and $D$ is the number of coordinates (2 in our case).
In addition, we perform several normalizations and augmentations explained in \Cref{sec:slr_library}.

\begin{figure}[h]
\includegraphics[width=\columnwidth]{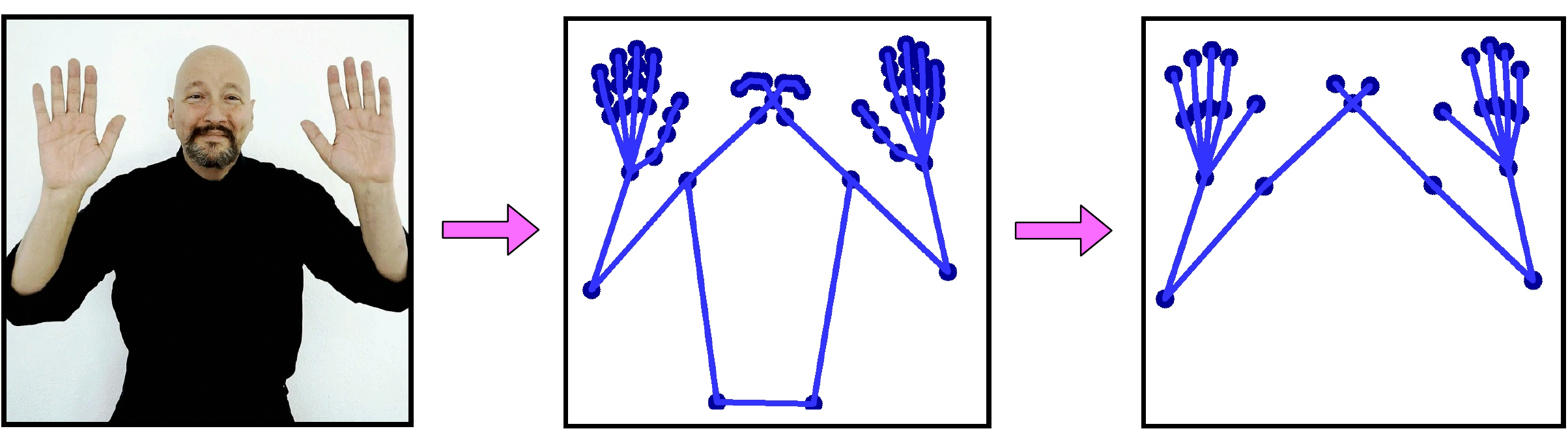}
\caption{Illustration for RGB frame to pose keypoints conversion. The center skeleton shows the upper portion of the 75 keypoints returned by MediaPipe, from which we choose only 27 points as shown in right.}
\label{figure:rgb_to_pose}
\end{figure}

\subsection{Standardized ISLR Models}

On the 7 different datasets we consider, different existing ISLR models have been trained which are detailed in \Cref{table:standardization_accuracies} which produce their current state-of-the-art results.
For INCLUDE dataset, an XGBoost model is used \cite{include_isl} with direct input as 135 pose-keypoints obtained using OpenPose.
For AUTSL, SL-GCN is used \cite{jiang2021skeleton} with 27 chosen keypoints as input from HRNet pose estimation model. 
For GSL, the corresponding model \cite{Parelli2020Exploiting3H} is an attention-based encoder-decoder with 3D hand pose and 2D body pose as input.
For WLASL, Temporal-GCN is used \cite{li2020word} by passing 55 chosen keypoints from OpenPose.
For LSA64, 33 chosen keypoints from OpenPose are used as input to an LSTM decoder \cite{Konstantinidis2018SIGNLR}.
For DEVISIGN, RGB features are used \cite{devisign_irdml} and the task is approached using a clustering-based classic technique called Iterative Reference Driven Metric Learning.
For CSL dataset, an I3D CNN is used as encoder with input as RGBD frames and BiLSTM as decoder \cite{Adaloglou_2021}.

The differences in the above models make it difficult to compare them on effectiveness, especially across diverse datasets. 
To enable standardized comparison of models, we train pose-based ISLR models on all datasets with similar training setups. 
These models belong to two groups: sequence-based models and graph-based models.
For sequence-based models we consider RNN and Transformer based architectures.
For the \textbf{RNN model}, we use a 4-layered bidirectional LSTM of hidden layer dimension 128 which takes as input the framewise pose-representation of 27 keypoints with 2 coordinates each, i.e., a vector of 54 points per frame.
We also use a temporal attention layer to weight the most effective frames for classification. 
For the \textbf{Transformer model}, we use a BERT-based architecture consisting of 5 Transformer-encoder layers with 6 attention heads and hidden dimension size 128, with a maximum sequence length of 256.
For the graph-based models we consider ST-GCN \cite{AAAI1817135} and SL-GCN \cite{jiang2021skeleton} models as discussed in \cref{sec:background}.
For \textbf{ST-GCN model}, we use 10 spatio-temporal GCN layers with spatial dimension of the graph consisting the 27 keypoints with a depth of 2 corresponding to the two coordinates.
For the \textbf{SL-GCN model}, we use again 10 SL-GCN blocks with the same graph structure and hyperparameters as the ST-GCN model.






\begin{table*}[h]
\begin{tabular}{ll|lr|rrrr} \toprule
\multicolumn{1}{c}{\multirow{2}{*}{\textbf{Dataset}}} & \multicolumn{1}{c}{\multirow{2}{*}{\textbf{Language}}} &  \multicolumn{2}{|c}{\textbf{State-of-the-art (pose) model}} & \multicolumn{4}{|c}{\textbf{Model available in \libfull}} \\
\multicolumn{1}{c}{}       &         & \textbf{Model (Params)}  & \textbf{Accuracy} & \textbf{LSTM} & \textbf{Transformer} & \textbf{ST-GCN} & \textbf{SL-GCN}         \\  \midrule

INCLUDE & Indian & Pose-XGBoost & 63.10 & 83.0 & 90.4 & 91.2 & \textbf{93.5} \\ 
AUTSL & Turkish & Pose-SL-GCN\footref{footnote:autsl_sota} (4.9M) & \textbf{95.02} & 77.4 & 81.0 & 90.4 & 91.9 \\ 
GSL & Greek & Pose-Attention (2.1M) & 83.42 & 86.6 & 89.5 & 93.5 & \textbf{95.4} \\ 
DEVISIGN\_L & Chinese & RGB-iRDML & 56.85 & 37.6 & 48.9 & 55.8 & \textbf{63.9} \\ 
CSL & Chinese & RGBD-I3D (27M) & 95.68 & 75.1 & 88.8 & 94.2 & \textbf{94.8} \\ 
LSA64 & Argentinian & Pose-LSTM (1.9M) & 93.91 & 90.2 & 92.5 & 94.7 & \textbf{97.8} \\
WLASL2000 & American & Pose-TGCN (5.2M) & 23.65 & 20.6 & 23.2 & 21.4 & \textbf{30.6} \\ \midrule
&& Average accuracy & $\rightarrow$ & 69.38 & 73.47 & 77.43 & 80.69 \\ \bottomrule
\end{tabular}
\caption{Accuracy of different models across datasets.}
\label{table:standardization_accuracies}
\end{table*}

\subsection{Experimental Setup and Results}\label{sec:benchmarking}

We train 4 models - LSTM, BERT, ST-GCN, and SL-GCN - for each of the 7 datasets. 
We use PyTorch Lightning to implement the data processing and training pipelines. We use Adam Optimizer to train all the models. 
For the LSTM model, we set the batch size as 32 and initial learning rate (LR) as $0.005$, while for BERT, we set a batch size 64, and LR of $0.0001$.
For ST-GCN and SL-GCN, we use a batch size of 32 and LR of 0.001.
We train all our models on a single NVIDIA Tesla V100 GPU.
Also for all datasets, we only train on the train-sets given, whereas most works (like AUTSL) train on both train-set and val-set to report the final test accuracy.
All trained models and the training configurations are open-sourced in \libfull.


\textbf{Accuracy} We report the obtained test-set accuracy of detecting individual signs, for each model against each dataset in \Cref{table:standardization_accuracies}. 
On all datasets, graph-based models report the state-of-the-art results. 
Except for AUTSL\footnote{SoTA AUTSL model is trained on very high quality pose data from HRNet pose estimator.\label{footnote:autsl_sota}}, on 6 of the 7 datasets, models we train improve upon the accuracy reported in the existing papers sometimes significantly (e.g., over 10\% on GSL).
These uniform results across a diverse set of SLs confirm that graph-based models on pose modality data define the SOTA.

\begin{table*}[h]
\centering
\begin{tabular}{l|rrrr|r}
\toprule
\textbf{Model} $\rightarrow$ & \textbf{LSTM} & \textbf{Transformer} & \textbf{ST-GCN} & \textbf{SL-GCN} & \textbf{SLDPC} \\
Params $\rightarrow$ & 1.6M & 3.8M & 2.9M & 4.9M & 4.0M \\
\midrule
\textbf{CPU} & \multicolumn{5}{c}{\textbf{Latency in milliseconds}} \\ \midrule
Xeon E5-2690 v4 (2.60GHz) & 08.05 & 30.64 & 23.02 & 52.8 & 47.60 \\
AMD Ryzen 7 3750H (2.30GHz) & 12.94 & 76.41 & 86.97 & 225.3 & 147.28 \\
Xeon Platinum 8168 (2.70GHz) & 05.38 & 23.76 & 51.64 & 112.66 & 112.52 \\
Xeon E5-2673 v4 (2.30GHz) & 09.03 & 43.69 & 99.39 & 201.31 & 188.43 \\
\bottomrule
\end{tabular}
\caption{Number of parameters and average latency of different model architectures}
\label{table:model_latencies}
\end{table*}

\textbf{Inference time}
Given that SLR is an interactive application, deployability atleast at 23 FPS without noticeable latency is essential.
We thus study the latency of our models on various CPU configurations so as to target ubiquitous deployment.
Details of the measurement setup and benchmarking of the pre-processing steps are in the \Cref{sec:inference_benchmarking}.
For each of the 4 models, we report the model size and latency measured on 4 different CPUs in \Cref{table:model_latencies}.
The LSTM model is an order of magnitude faster across all devices than the most accurate SL-GCN model, and is a good candidate when speed is essential at the cost of about 10\% accuracy drop that we observed in \Cref{table:standardization_accuracies}.
Amongst the graph-based methods, ST-GCN provides a good trade-off being about 2$\times$ faster than SL-GCN at the cost of only 3\% lower average accuracy across datasets. 

In summary, the standardized benchmarking of multiple models in terms of accuracy on datasets and latency on devices informs model selection.
Making the trade-off between accuracy and latency, we use the ST-GCN model for the pretrained model we discuss later. 
Our choice is also informed by the cost of the training step: The more accurate SL-GCN model takes 4$\times$ longer to train than ST-GCN.

\section{Self-Supervised Learning for ISLR}\label{sec:ssl_islr}


In this section, we describe our efforts in building the largest corpus for self-supervised pretraining and our experiments in different pretraining strategies. 

\subsection{Indian SL Corpus for Self-supervised pretraining}

\begin{table}[h!]
\centering
\begin{tabular}{lrll}
\toprule
\textbf{Channel} & \textbf{Hours} & \textbf{Domain} & \textbf{Duration} \\ \midrule
NewsHook & 615 & News & 3-4mins \\ 
MBM Vadodara & 225 & News & 7-8mins \\ 
ISH-News & 145 & News & 3-5mins \\
NIOS & 115 & Educational & 2-30mins \\ 
SIGN Library & 29 & Educational & 5-10mins \\ \midrule
\textbf{Total} & \textbf{1129} & & \\ \bottomrule
\end{tabular}
\caption{Source-wise statistics of the processed self-supervised dataset on Indian-SL}
\label{table:youtube_channels}
\end{table}

Large text corpora such as BookCorpus, Wikipedia dumps, OSCAR, etc.~have enabled pretraining of large language models in NLP.
Although there are large amounts of raw sign language videos available on the internet, no existing work has studied how such large volumes of open unlabelled data can be collected and used for SLR tasks. 
To address this, we create a corpus of Indian SL data by curating videos, pre-process the videos, and release a standardized pose-based dataset compatible with the models discussed in the previous section.

We manually search for freely available major sources of Indian SL videos. 
We restrict our search to a single sign language so as to study the effect of pretraining on same language and crosslingual ISLR tasks.
We sort the sources by the number of hours of videos and choose the top 5 sources for download.
All of these 5 sources, as listed in \Cref{table:youtube_channels} are YouTube channels, totalling over 1,500 hours.
We downloaded these videos resulting in an uncompressed dataset of size 1.1 TB. 
This is done on a large machine with 96 CPU cores in a multi-threaded approach, which took around 3 days to completely crawl all the mentioned sources.

We only chose YouTube channels whose content license is \textit{Creative Commons} and ensured that significant number of videos only have a single signer.
Our video sources are from news and educational domains. Around 87\% of the total data is from news channels.
The Education domain channels are \textit{National Institute of Open Schooling} (NIOS), an intiative by Government of India and SIGN Library channel, an initiative to make educational content in Indian SL.

\begin{figure}[h!]
\includegraphics[width=\columnwidth]{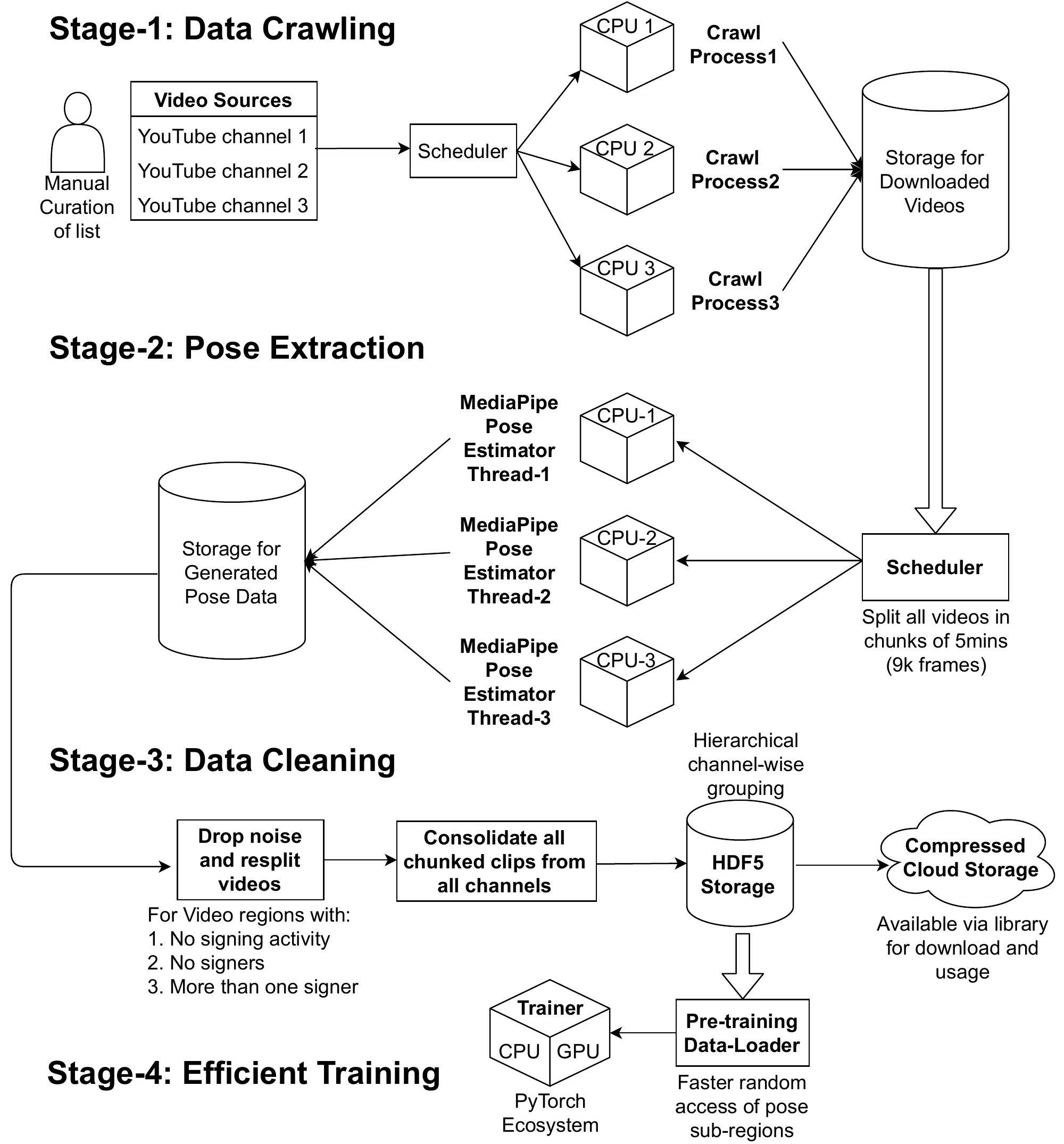}
\caption{Pipeline used to collect and process Indian SL corpus for self-supervised pretraining}
\label{fig:data_processing_pipeline}
\end{figure}

We pass these downloaded videos through a processing pipeline as described in \Cref{fig:data_processing_pipeline}.
We initially dump the pose data for all videos, then process them to remove those which are noisy or contain either no person or more than 1 person. 
This resulted in 1,129 hours of Indian SL data, as detailed source-wise in \Cref{table:youtube_channels}.
This is significantly larger than all the training sets in the datasets we studied which is on average 177 hours.
We pass these videos through MediaPipe to obtain pose information as described earlier, i.e., 75 keypoints per frame.
The resultant Indian SL corpus has more than 100 million pose frames.
We convert this to the HDF5 format to enable efficient random access, as is required for training.
We open-source this corpus of about 250 GB which is available in \libfull.



\subsection{Pretraining Setup and Experiments}\label{sec:ssl_experiments}

We explore the three major pretraining strategies as described in \Cref{sec:pretraining_strategies} and explain how and why certain self-supervised settings are effective for ISLR.
We pretrain on randomly sampled consecutive input sequences of length 60-120 frames (approximating 2-4 secs with 30fps videos).
After pretraining, we fine-tune the models on the respective ISLR dataset with an added classification head.

\begin{table}[h!]
\centering
\begin{tabular}{ll}
\toprule
\textbf{Training of ST-GCN} & \textbf{Accuracy} \\ \toprule
No pretraining + Fine-tune & 91.2 \\ \midrule
Masked-based + Fine-tune & 91.3 \\ 
Contrastive learning + Fine-tune & 90.8 \\ 
Predictive-coding + Fine-tune & 94.7 \\
\bottomrule
\end{tabular}
\caption{Effectiveness of pretraining strategies as measured on ISLR accuracy on INCLUDE}
\label{table:pretraining_accuracies}
\end{table}

\subsubsection{Masking-based pretraining}
\begin{figure}[h]
\includegraphics[width=\columnwidth]{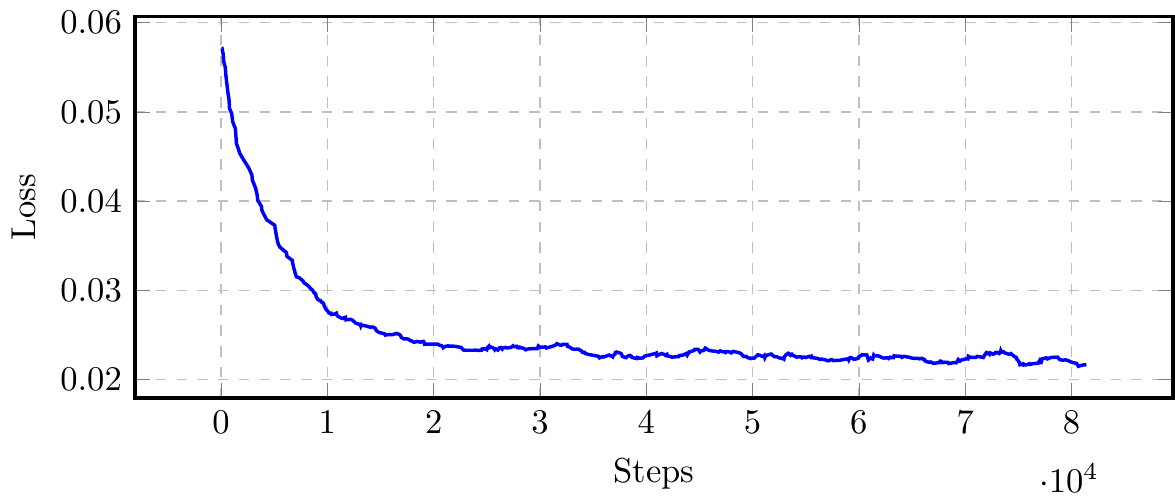}
\caption{Loss curve for masked pretraining with regression loss}
\label{figure:mlm_pretraining_loss}
\end{figure}

We follow the same hyperparameter settings as described in Motion-Transformer \cite{motion_transformer}, to pretrain a BERT-based model with random masking of 40\% of the input frames.
When using only the regression loss, we find that pretraining learns to reduce the loss as shown in \Cref{figure:mlm_pretraining_loss}.
However, when fine-tuned on the INCLUDE dataset, we see no major contribution of the pretrained model to increasing the accuracy as shown in  \Cref{table:pretraining_accuracies}.
We posit that while pretraining was able to approximate interpolation for the masked frames based on the surrounding context, it did not learn higher-order features relevant across individual signs.
We also experiment with different masking ratios (20\% and 30\%) as well as different length of random contiguous masking spans (randomly selected between 2-10), and obtain similar results.

\begin{figure}[h]
\includegraphics[width=\columnwidth]{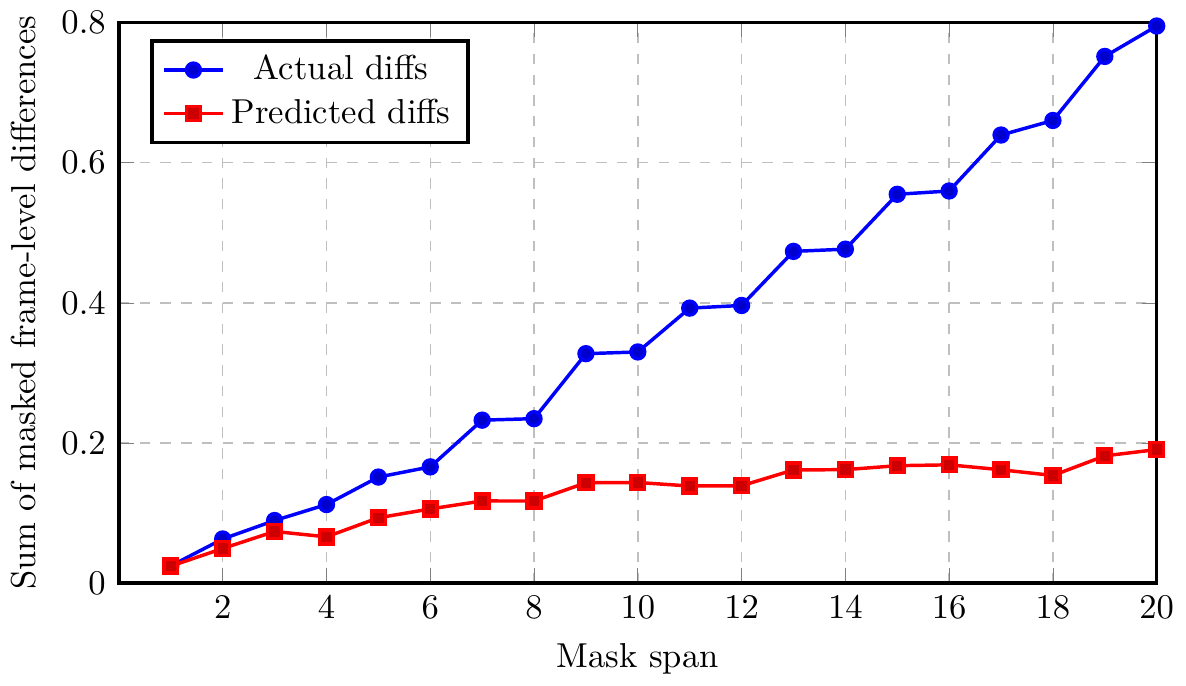}
\caption{Differences in the output range of masked predictions of pretrained model and corresponding actual keypoints}
\label{figure:mlm_pred_diff}
\end{figure}

To explain this behaviour, we analyzed the input data as well as the outputs by the model.
We find that the model was able to converge because learning to perform an approximate linear interpolation for the masked frames based on the surrounding context was sufficient reduce the loss significantly.
However, we posit that such interpolation does not learn any high-level features.
This is illustrated in \Cref{figure:mlm_pred_diff}, where for each masking span length, we plot the sum of absolute differences between each consecutive masked frames $F_i$ and $F_{i-1}$, for both predictions from the model as well as the actual frame keypoints.
The numbers shown are averaged across all videos in the INCLUDE test set, in which the masking is done around the center region of each video.
The plot shows that as masking length is increased, the gap between the predicted values and the actual values diverges indicating an inability to learn longer-range patterns that may be necessary to classify signs.

\begin{figure}[h]
\includegraphics[width=\columnwidth]{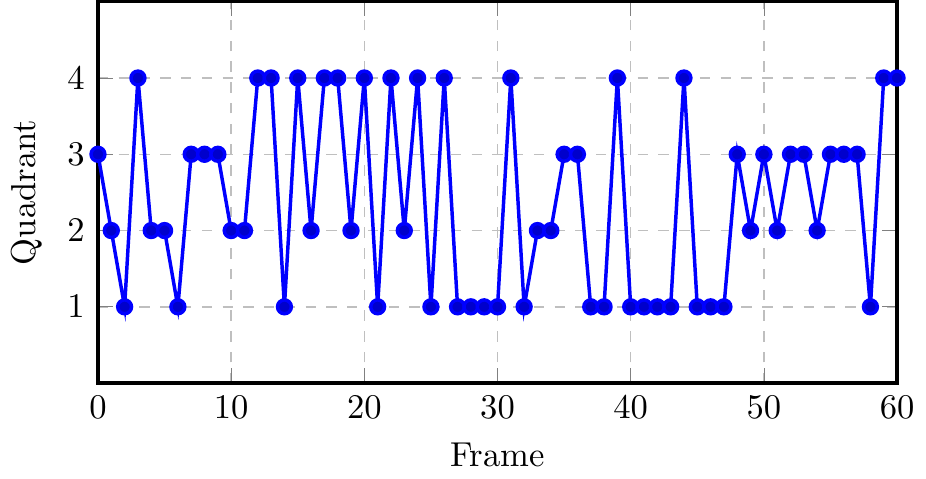}
\caption{Sample visualization of direction labels for keypoint-15 from the frames of a random INCLUDE video (\textit{Adjectives/4. sad/MVI\_9720})}
\label{figure:mlm_direction_targets}
\end{figure}

We also experiment with pretraining using direction loss as explained in background, which essentially is an objective to classify which quadrant the motion vector for each frame will lie.
We find that the pretraining does not converge. 
Upon checking the labels, we see that at the fine-grained level of each frames, the approximately discretized quadrant for each motion vector were seemingly almost random because of the slightly jittery predictions for each frame by the pose estimation model. 
Also, since the quadrant-type classification encodes only 4 directions, it fails to capture static motion (keypoints which do not move much temporally), which accounts for more than half of the total motion vectors.
We thus posit that the direction classification targets are noisy and do not allow the pretraining loss to converge.
\Cref{figure:mlm_direction_targets} shows the visualization of quadrants for a randomly-selected joint from a random video in the INCLUDE dataset, to visually verify how noisy the targets for direction loss are.

We leave it to future works to study specially designed SL-domain specific abstract representations of pose, which might try to solve the issue of modelling the outputs of BERT as abstract latent representations instead of directly posing an interpolation-like task for the masked tokens (generally achieved by having specific encoder and decoder around BERT).

\subsubsection{Contrastive-learning based}
\begin{figure}[h]
\includegraphics[width=\columnwidth]{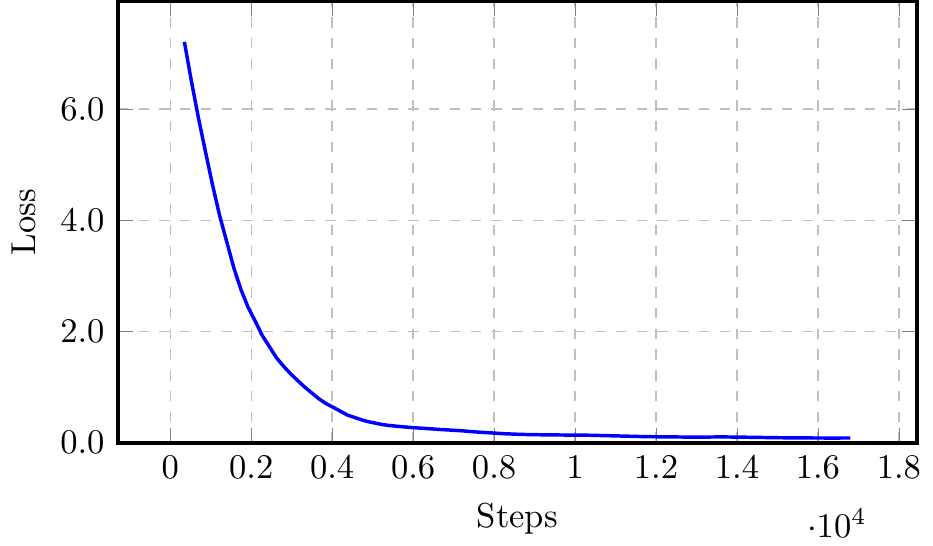}
\caption{Loss curve for contrastive pretraining}
\label{figure:cl_pretraining_loss}
\end{figure}

Inspired by \cite{contrastive_gao21a}, we consider Shear, Scaling and Rotation augmentations for each frame, and pretrain the model.
For pretraining, we used a batch size of 128 and for finetuning, we used a batch size of 64.
For both pretraining and finetuning, we used Adam optimizer with an initial learning rate of 1e-3.
To obtain negative samples, we use a Memory Bank to obtain the embeddings from samples of recent previous batches, which is essentially a FIFO queue of fixed size.
We use Facebook's \textit{MoCo} code\footnote{\url{https://github.com/facebookresearch/moco}} to implement the contrastive learning setup, by plugging-in our ST-GCN as the encoder.
We observe that it converges on reducing the InfoNCE loss (as seen in \Cref{figure:cl_pretraining_loss}).

\begin{figure}[h]
\includegraphics[width=\columnwidth]{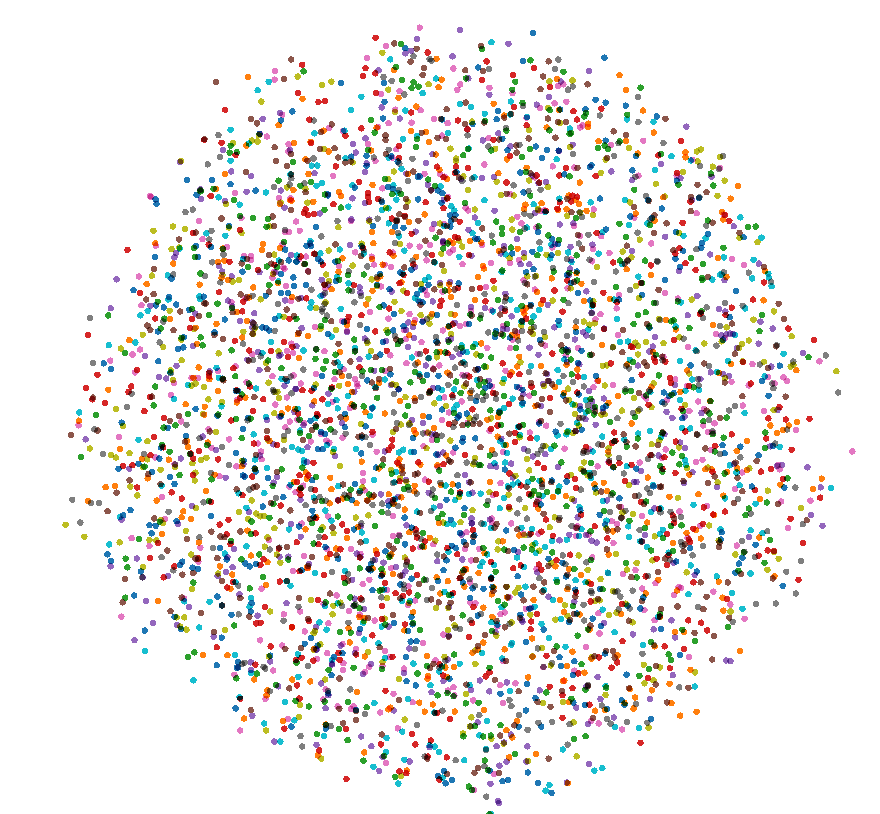}
\caption{PCA visualization of INCLUDE50 embeddings obtained from Contrastive-Learning pretrained model}
\label{figure:cl_embedding_visualization}
\end{figure}

We then fine-tune on INCLUDE and again did not observe any gain over the baseline of training from scratch as seen in \Cref{table:pretraining_accuracies}.
That is, although the pretraining converges, the representations learnt do not signify any semantic relationships in the signs.
To illustrate this, we take a standard subset of the INCLUDE dataset, called INCLUDE50 (containing 50 classes) and visualize the embeddings of all signs using PCA clustering.
Note that each class is uniquely colored to identify if similar signs are grouped together.
\Cref{figure:cl_embedding_visualization} shows that the learnt embeddings do not discriminate the classes, suggesting that the embeddings may not be informative for the downstream sign recognition task.
In conclusion, using the embeddings of data from the pretrained model, we observed two facts: (a) Embeddings of different augmentations of a video clip are similar indicating successful pretraining, but (b) Embeddings of different videos from the INCLUDE dataset do not show any clustering based on the class.
Hence, we posit that pretraining did not learn higher order semantics that could be helpful for ISLR.

\subsubsection{Predictive-coding based} 
\begin{figure}[h]
\includegraphics[width=\columnwidth]{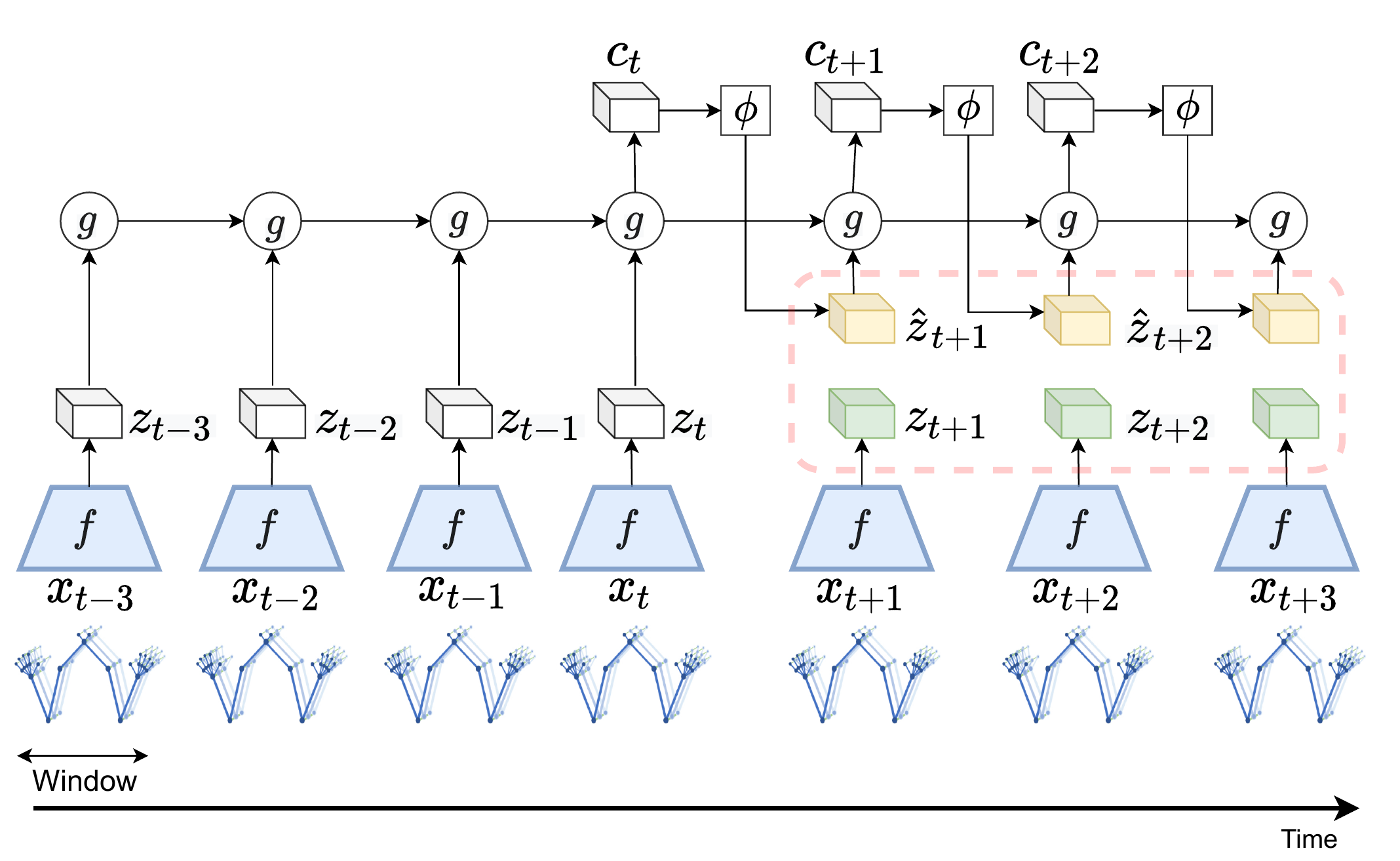}
\caption{Model architecture for DPC pretraining}
\label{figure:dpc_architecture}
\end{figure}

Our architecture is inspired from Dense Predictive Coding \cite{han2019video}, but using pose modality.
The architecture is represented in \Cref{figure:dpc_architecture}.
The pose frames from a video clip will be partitioned into multiple non-overlapping windows with equal number of frames in each window.
The encoder $f$ takes each window of pose keypoints as input and embeds into the hidden space $z$.
Specifically, the ST-GCN encoder embeds each input window $x_i$, and the direct output is average pooled across the spatial and temporal dimensions to obtain the output embedding $z_i$ for each window.
The embeddings are then fed to a Gated Recurrent Unit (GRU) as a temporal sequence and the future timesteps $\hat{z}_{i}$ are predicted sequentially using the past timestep representations from GRU, with an affine transform layer $\phi$.
We use 4 windows of data as input to predict the embeddings of the next 3 windows, each window spanning 10 frames, which we empirically found to be the best setting.
For pretraining, we used a batch size of 128 and for finetuning, we used a batch size of 64.
For both pretraining and finetuning, we used Adam optimizer with an initial learning rate of 1e-3.

\begin{figure}[h]
\includegraphics[width=\columnwidth]{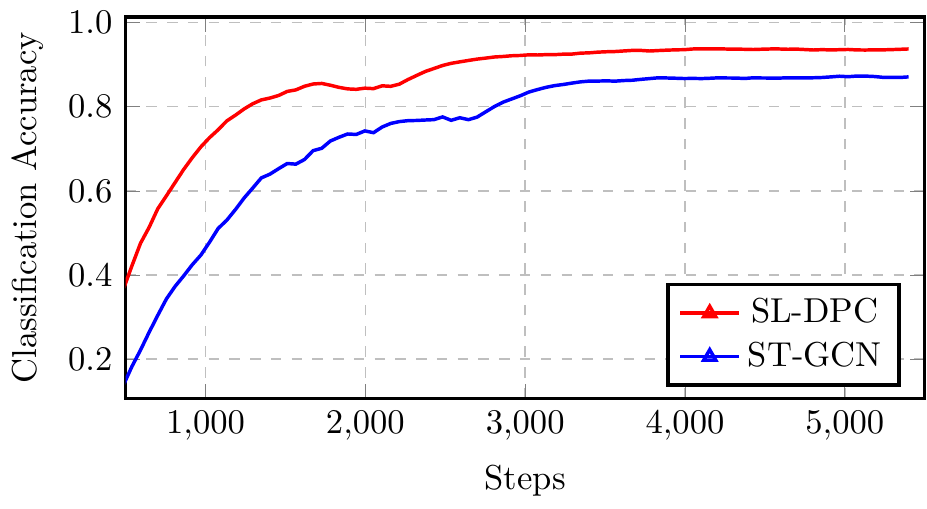}
\caption{DPC Fine-tuning (orange) vs fresh training (light-green) validation accuracy plot}
\label{figure:pretraining_accuracy_plot}
\end{figure}

Upon fine-tuning on INCLUDE, DPC provides a significant improvement of 3.5\% over the baseline.
\Cref{figure:pretraining_accuracy_plot} shows the the validation accuracy between baseline and finetuned model, indicating the performance gap between fine-tuning and an ST-GCN model being trained from scratch.
We posit that DPC is successful, while previous methods were not, as it learns coarse-grained representations across multiple frames and thereby captures motion semantics of actions in SL.
This clearly demonstrates that self-supervised learning produces a significant boost in performance for downstream tasks.

All pretrained models and scripts are open-sourced through \libfull.
To the best of our knowledge, this is the first comparison of pretraining strategies for SLR.

\subsection{Evaluation on low-resource and crosslingual settings}

We demonstrated that DPC-based pretraining is effective. 
We now analyze the effectiveness of such pretraining in two constrained settings - (a) when fine-tuning datasets are small, and (b) when fine-tuning on sign languages different from the sign language used for pretraining.
The former captures in-language generalization while the latter crosslingual generalization.


\subsubsection{In-language generalization}

The INCLUDE dataset contains an average of 17 samples per class.
For this setting, we observed a gain of 3.5\% with DPC-based pretraining over training from scratch.
How does this performance boost change when we have fewer samples per class?
We present results for 10, 5, and 3 samples per class in \Cref{table:finetuning_accuracies}.
We observe that as the number of labels decreases the performance boost due to pretraining is higher indicating effective in-language generalization.


\begin{table}[htp]
\begin{tabular}{llll}
\toprule
\textbf{Dataset} & \textbf{Samples/class} & \textbf{ST-GCN} & \textbf{DPC} \\ \midrule
\multirow{4}{*}{\shortstack{INCLUDE\\ (Indian)}} & Full (Avg. 17) & 91.2 & 94.7 \\
                         & 10 & 79.7 & 86.27 \\
                         & 5 & 45 & 57.35 \\
                         & 3 & 15.2  & 35.42 \\ \midrule
\multirow{3}{*}{\shortstack{WLASL2000\\ (American)}} & Full (Avg. 10) & 21.4 & 27.4 \\
                           & 5 & 3.1 & 5.74 \\
                           & 3 & 1.6 & 2.78 \\ \midrule
\multirow{3}{*}{\shortstack{DEVISIGN\_L\\ (Chinese)}} & Full (8) & 55.8 & 59.5 \\
                           & 5 & 33.0 & 40.26 \\
                           & 3 & 8.46 & 18.65 \\ \midrule
\multirow{3}{*}{\shortstack{LSA64\\ (Argentinian)}} & Full (50) & 94.7 & 96.25 \\
                           & 5 & 64.7 & 75.32 \\
                           & 3 & 39.7 & 57.19 \\ \bottomrule
\end{tabular}
\caption{Effectiveness of pretraining for in-language (first row) and crosslingual transfer (last three rows)}
\label{table:finetuning_accuracies}
\end{table}

\subsubsection{Crosslingual transfer}

Does the pretraining on Indian sign language provide a performance boost when fine-tuning on other sign languages?
We study this for 3 different sign languages - American, Chinese, and Argentinian - and report results in \Cref{table:finetuning_accuracies}.
We see that crosslingual transfer is effective leading to gains of about 6\%, 4\%, and 2\% on the three datasets, similar to the 3\% gain on in-language accuracy.
Further, we also observe that these gains extend to low-resource settings of fewer labels per sign.
For instance on Argentinian SL, with 3 labels, pretraining on Indian SL given an improvement of about 18\% in accuracy.
To the best of our knowledge this is the first successful demonstration of crosslingual transfer in ISLR.

In summary, we discussed different pretraining strategies and found that only DPC learns semantically relevant higher-order features.
With DPC-based pretraining we demonstrated both in-language and crosslingual transfer.



\section{The \libname library}\label{sec:slr_library}

As mentioned in the previous sections, we open-source all our contributions through the \libfull~library.
This includes the pose-based datasets for the 6 SLs, 4 ISLR models trained on 7 datasets, the pretraining corpus on Indian SL with over 1,100 hours of pose data, pretrained models on this corpus for all 3 pretraining strategies, and models fine-tuned for 4 different SLs on top of the pretrained model.
We also provide scripts for efficient deployment using MediaPipe pose estimation and our trained ISLR models.


We encourage researchers to contribute datasets, models, and other utilities to make sign language research more accessible. 
We are particularly interested to support lesser studied and low-resource SLs from across the world. 


\subsection{Inference Benchmarking}\label{sec:inference_benchmarking}

In this section, we explain how we achieve over 23fps real-time inference, by using MediaPipe Holistic for generating poses (as an ISLR encoder) and our pose-based models (as decoder) that recognizes the sign at any given window.

\subsubsection{MediaPipe Inference}

For pose-estimation, MediaPipe offers 3 variants of models: \textit{heavy}, \textit{full} and \textit{lite} in decreasing order of accuracy but increasing order of inference-speed.
The latency of these variants on Intel Xeon E5-2690 v4 CPU with a frame-size of 640x480 were 142.59ms, 55.28ms, and 35.37ms respectively per frame.
For all training and testing in this work, we used the heavy model to get the best quality results.

For real-time inference, depending on one's CPU, either of the 3 variants can be used with the trained models, since all the 3 BlazePose models are trained on the same dataset to return same number of keypoints.
Based on our experience, we prefer only \textit{lite} or \textit{full} variants depending on the CPU-type, and we find the \textit{heavy} model only suitable if we employ frame-skipping and use decoder models that also work at a lower FPS (below 8fps).
We leave study of low-fps ISLR models open for future research.

\subsubsection{ISLR Model Inference}

The benchmarking is done with a batch size of 1 with complete serial processing (without any data loading parallelization).
The latencies reported in the table corresponds to average inference time per video using the test set of the INCLUDE dataset, for both freshly trained models and pretrained sign language DPC (SLDPC) model.

Note that encoder (pose estimation) and decoder (classifier) are parallelized such that the former is a producer of skeletons for window of live frames, and the latter is a consumer which recognizes glosses.

\subsection{Pose Transforms}\label{sec:pose_transforms}
The library provides utilities that are helpful specifically for processing pose-based data during training and inference. Currently, the following data normalization and augmentation techniques are supported:



\subsubsection{Pre-processing}
Normalization is generally done to convert all the data to a form that is invariant to many attributes including noise. The most common pre-processing techniques are described below:

1. \textit{VideoDimensionsNormalize}: Different videos would be in different resolutions, hence the scale of pose data from different videos would be variant. Generally pose keypoints are normalized by dividing all coordinates $(x,y)$ by the width and height of each video/frame.

2. \textit{CenterAndScaleNormalize}: To normalize all of the pose data to be in the same scale and reference coordinate system, we can normalize every pose by a constant feature of their body. For example, we can use the average span of the person's shoulders or spinal cord throughout the video to be a constant width and place the center of the span at $(0,0)$ in the plane. This is inspired from a python library called \textit{pose-format}\footnote{\url{https://github.com/AmitMY/pose-format}}.

3. \textit{PoseInterpolation}: Generally, keypoint estimation models are run for each frame in a video. If some of the frames in a video are blurred, there is high chance that pose generation would fail for those frames. To handle such noise in the data, for all frames which have their keypoints corrupt or missing, we fill them up by interpolation of the adjacent frames to the left and right. This is also essential when there is a minimum number of frames required for each instance, but certain video clips have lower number of frames than required.

\subsubsection{Augmentations}
Owing to very small sizes of ISLR datasets, the data is generally not as robust to cover all cases as in real-time scenarios. Hence it is essential to augment the data to account for different variations that are not represented in the training distribution including noise. Some of the important augmentations are described below briefly:

1. \textit{ShearTransform}: It is used to displace the joints in a random direction. The shear matrix for 2D can be given by:
\begin{equation}
\small
    \mathbf{S}=\left[
    \begin{array}{cc}
    1   &   s_{x} \\
    0  &   1  \\
    \end{array}
    \right]
\end{equation}

The $s_x$ value is a randomly sampled shear vector. Multiplying $S$ with the the actual coordinates to get the new coordinates.

2. \textit{RotatationTransform}: We can simulate the viewpoint changes of the camera by using this rotation augmentation. The standard rotation matrix for 2D can be given by:

\begin{equation}
\small
    \mathbf{R}=\left[
    \begin{array}{cc}
    \cos \theta &-\sin \theta \\
    \sin \theta &\cos \theta  \\
    \end{array}
    \right]
\end{equation}

We select a random rotation angle from -$\pi$/3 to $\pi$/3 and we multiply the matrix R with the actual coordinates matrix to get the rotated matrix.

3. \textit{ScaleTransform}: This is used to simulate different scales of the pose data to account for relative zoomed-in or zoomed-out view of signers from the camera. A random number is sampled and multiplied with the coordinates for each video. This is generally not necessary when the \textit{CenterAndScaleNormalize} is done.

4. \textit{PoseRandomShift}: This is used to move a significant portion of the video by a time offset $T_{offset}$ so as to make the ISLR models robust to inaccurate segmentation of real-time video. This can also be used to randomly distribute the zero padding at the end of a video to initial and final positions.

5. \textit{UniformTemporalSubsample}: In cases where the number of frames in a video clip exceeds a maximum limit, it maybe useful to uniformly sample frames from the video instead of considering only the initial $N_{max}$ frames. This is also referred as \textit{FrameSkipping} if the problem is explicitly posed in-terms of number of frames once the sample is to be taken.

6. \textit{RandomTemporalSubsample}: Instead of always sticking to \textit{UniformTemporalSubsampling} for the above case, it is also a good augmentation to sample a random fixed contiguous window of required size covering a maximum number of frames. This has an effect similar to \textit{PoseRandomShift}.

\subsection{Library Structure}

This subsection briefly describes how the library is modularized in such a way that it is easily extensible for different purposes.
Any important task like training, testing or inference can be easily run just using a config file which can be passed to the toolkit, which will take care of the end-to-end processing; hence any beginner can easily get started with SLR research.
Moreover, addition of more modules or features is easy and beginner-friendly; hence not compromising on flexibility.

Any model in the library is abstracted as an encoder-decoder model. This ensures that there is no redundancy at any level.
For example, for the ST-GCN model, the encoder is an instance of ST-GCN module and decoder is an instance of fully-connected classifier.
Moreoever, if one wants to train a new model like BERT-GCN \cite{lin2021bertgcn}, one can easily mention the encoder as ST-GCN and decoder as BERT, and use the library directly without any changes;
or extend the library to add support for any new encoder (like GPT) or decoder (feature pooling).

Any dataset supported in the library is extended from a base dataset class, which handles most of the common processing.
Hence each dataset class is only required to mention how to read the labels and training data for the specific datasets.
This makes it easier to extend the library for any new dataset just with a few lines of code.

All the aspects of the toolkit are well-documented online\footnote{\url{https://openhands.readthedocs.io}} for anyone to get started easily.
The library is fully python-based.
Beginners can directly utilize the toolkit using configs without writing any code, and researchers can import any module from the library into their code and customize it as required for their pipeline.

\section{Conclusion and Future work}\label{sec:future_works}

In this work, we make several contributions to make sign language research more accessible. 
We release pose-based datasets and 4 different ISLR models across 6 sign languages.
This evaluation enabled us to identify graph-based methods such as ST-GCN as being accurate and efficient.
We release the first large corpus of SL data for self-supervised pretraining.
We evaluated different pretraining strategies and found DPC as being effective. 
We also show that pretraining is effective both for in-language and crosslingual transfer.
All our models, datasets, training and deployment scripts are open-sourced in \libfull.

Several directions for future work emerge such as 
using face landmarks along with the current keypoints,
better high quality pose-estimation models like MMPose\footnote{\url{https://mmpose.readthedocs.io}},
evaluating alternative graph-based models, 
efficiently sampling the data from the raw dataset such that the samples are diverse enough,
and quantized inference for 2$\times$-4$\times$ reduced latency.
On the library front, we aim to release updated versions incorporating more SL datasets, better graph-based models, studying the performance on low FPS videos (like 2-4 FPS), effect of pretraining using other high-resource SL datasets, extending to CSLR, and improving deployment features.


\section*{Acknowledgements}

We would like to thank \textit{Aravint Annamalai} from IIT Madras for preparing the list of potential YouTube channels that can be crawled and for his help in downloading them.
We would like to thank the entire \textit{AI4Bharat Sign Language Team}\footnote{\url{https://sign-language.ai4bharat.org}} for their support and feedback for this work, especially from \textit{Rohith Gandhi Ganesan} for his insights on code structuring, and \textit{Advaith Sridhar} for managing the overall project.
We would also like to extend our immense gratitude to Microsoft's \textit{AI for Accessibility} program for granting us the compute required to carry out all the experiments in this work, through Microsoft Azure cloud platform.
Our extended gratitude also goes to Zenodo, who helped us with hosting our large datasets \cite{gokul_nc_2021_5558897}.
Finally, we thank all the content creators and ISLR dataset curators without whose data this work would have been impossible.

\newpage

\bibliographystyle{aaai}
\bibliography{main}




\end{document}